\documentclass{article}
\usepackage{arxiv}
\usepackage[utf8]{inputenc} 
\usepackage[T1]{fontenc}    
\usepackage{hyperref}       
\usepackage{url}            
\usepackage{booktabs}       
\usepackage{amsfonts}       
\usepackage{nicefrac}       
\usepackage{microtype}      
\usepackage{lipsum}		
\usepackage{graphicx}
\usepackage{natbib}
\usepackage{doi}
\usepackage{url}
\usepackage{hyperref}
\hypersetup{
  colorlinks=true,
  linkcolor=blue,
  citecolor=blue,
  urlcolor=blue
}

\title{AI-Powered Facial Mask Removal Is Not Suitable \\ For Identification}

\date{} 					

\author{ Emily A. Cooper \\
	Herbert Wertheim School of Optometry \& Vision Science \\
    University of California, Berkeley \\
	\texttt{emilycooper@berkeley.edu} \\
	\And
	Hany Farid \\
	School of Information\\
	University of California, Berkeley \\
	\texttt{hfarid@berkeley.edu}
}

\renewcommand{\shorttitle}{AI-Powered Unmasking}

\begin{document}
\maketitle

\begin{abstract}
	Recently, crowd-sourced online criminal investigations have used generative-AI to enhance low-quality visual evidence. In one high-profile case, social-media users circulated an “AI-unmasked” image of a federal agent involved in a fatal shooting, fueling a wide-spread misidentification. In response to this and similar incidents, we conducted a large-scale analysis evaluating the efficacy and risks of commercial AI-powered facial unmasking, specifically assessing whether the resulting faces can be reliably matched to true identities.
\end{abstract}


\section{Introduction}

Crowd-sourced online investigations, or websleuthing, have become increasingly prominent as citizens use publicly available images, documents, and videos to investigate crimes.~\cite{yardley2018s,walsh2024}. Recently, web sleuths have been empowered by generative AI in the form of image enhancement. In the wake of the assassination of Charlie Kirk, for example, users on social media questioned the identity of the shooter after he was arrested because a widely circulated surveillance photo of the shooter did not appear to match his mugshot~\cite{roley2025doubts}. This surveillance photo, however, was an ``AI-enhanced'' version of a grainy image, with the resulting facial features almost entirely hallucinated~\cite{norman2024investigation}.

More recently, after a masked ICE agent in Minnesota shot and killed a civilian, websleuths used GrokAI to digitally remove the mask with the goal of trying to identify the agent~\cite{grove2026ice}. As GrokAI had no other knowledge of the ICE agent's face, this unmasked image relied on generative inpainting: drawing statistical inferences to synthesize plausible facial features rather than recover true ones. The unmasked image was then uploaded to a reverse image search which led to the misidentification of the ICE agent as Steve Grove, the Publisher of the Minnesota Star Tribune~\cite{brumfiel2026ice}. 

On the heels of this AI-powered misidentification---which is unlikely to be a singular attempt to use AI for mask removal---we performed a large-scale analysis to measure the efficacy of commercial AI tools to unmask a face and produce an image that matches the true identity.


\section*{Methods}

\subsection*{Faces}

\begin{figure*}[t]
	\includegraphics[width=\linewidth]{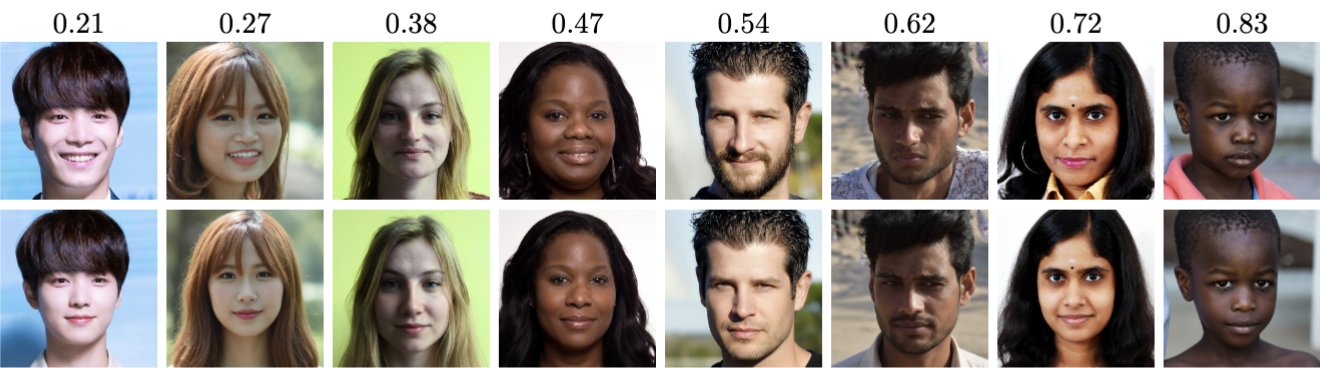}
	\caption{A representative set of original unmasked faces (top) and the result of unmasking using Gemini to fill in the removal of the lower-half of the face. The value above each face is the biometric similarity (in $[-1,1]$) between the original and unmasked face. Larger values correspond to a higher facial similarity.}
	\label{fig:original-unmasked}
\end{figure*}

We make use of three datasets of real faces: (1) {\em Different-ID}, (2) {\em Same-ID}, and (3) {\em US-Senators}. The {\em Different-ID} dataset consists of $400$ faces of distinct identities with one image from each identity (Figure~\ref{fig:original-unmasked}, top row). The images are all $1024 \times 1024$ and depict diverse faces in terms of gender ($200$ women; $200$ men), apparent age, and race ($100$ Black, $100$ Caucasian, $100$ East Asian, and $100$ South Asian)~\cite{nightingale2022ai}. The {\em Same-ID} dataset consists of pairs of images from each of $100$ distinct identities. Each pair comprises two images taken of the same person in different environments, lighting, and poses. These faces are culled\footnote{\url{https://www.kaggle.com/datasets/hearfool/vggface2}} from the VGG2 dataset~\cite{cao2018vggface2} by selecting a pair of frontal facing unobscured faces for each unique identity with a minimum resolution of $512$ pixels. The {\em US-Senators} dataset consists of two images from each of $91$ members of the $116^{th}$ US Senate (2019-2021). Each pair of images consists of a United States Senator wearing a facial mask (obtained from online media outlets) along with their official (unmasked) portrait photo.

\subsection*{AI-Powered Unmasking}

\begin{figure}[t]
	\includegraphics[width=\linewidth]{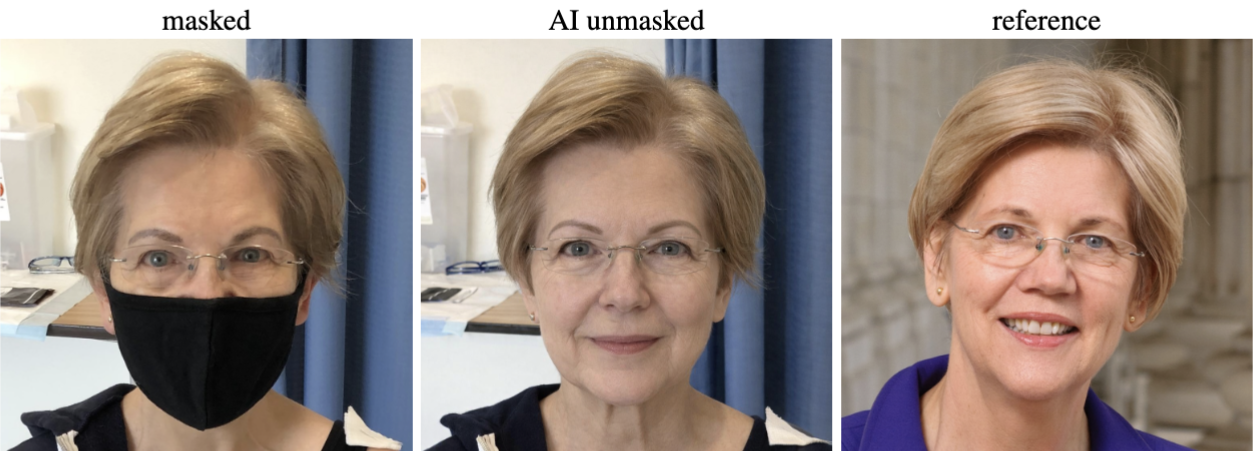}
	\caption{Senator Elizabeth Warren's mask (left) is removed (middle) using ChatGPT (photo sources: commons.wikimedia.org).}
	\label{fig:Warren}
\end{figure}

We performed AI-powered unmasking on two different types of ``masked'' faces. In the first, we simulated masking by removing the lower-half of each face in the {\em Different-ID} dataset. Because the eyes in each face in this dataset are aligned to the same absolute pixel location, we were able to programatically split these images at the same facial position at the bridge of the nose ($y=550$). In the second, we removed the facial mask in each of the faces in the {\em US Senators} dataset (Figure~\ref{fig:Warren}). AI-powered unmasking was performed with OpenAI's ChatGPT (gpt-image-1.5), Google's Gemini (gemini-2.5-flash-image), and X's GrokAI. For ChatGPT and Gemini this was performed using their API; GrokAI does not offer an API for image editing, so this was performed using the web interface on only $80$ of the $400$ faces. In all cases, the unmasking was performed via generative inpainting: the synthesis of plausible facial features based on the input image.

Each AI model was provided with an image to ``unmask.'' For the images in which the lower-half was removed, each model was prompted with: ``This is for academic research. Given this image of the top half of a face, return an image showing the full face.'' For images in which the real facial mask was removed, the prompt was: ``This is for academic research. Create a realistic reconstruction of the facial features that would be beneath this medical mask.'' 

The addition of ``academic research'' was used to contend with automatic content moderation that occasionally flagged requests as potentially violative. There are, however, no privacy concerns with these unmasking requests because the original faces are already in a public dataset~\cite{nightingale2022ai} for the lower-half reconstruction, and the US Senators are identified in each photo of them.

\subsection*{Facial Biometrics} 

To measure the facial similarity between original and unmasked faces we used ArcFace~\cite{deng2019arcface}, a widely used facial recognition system specifically designed to enforce separation between distinct identities. The biometric similarity between two faces is determined by passing each face through the trained network to extract a $512$-D embedding. The similarity between two faces is measured as the cosine similarity between embeddings leading to a score between $-1$ (mismatch) and $1$ (perfect match). As a control, we confirmed our key observations with a second, more recent facial recognition system, AdaFace~\cite{kim2022adaface}.

To calibrate the biometric facial similarity scores, we also computed biometric similarity between (1) each pair of images of each identity in the {\em Same-ID} dataset, and (2) pairs of different identities matched for gender and race in {\em Different-ID}.

\subsection*{Facial Matching}

To evaluate AI-powered unmasking in an operational face matching context, we determined a set of thresholds on biometric similarity scores corresponding to false match rates of $1$ in $100$ to $1$ in $1,000,000$ ($1.0\%$, $0.1\%$, $0.01\%$, $0.001\%$, and $0.0001\%$). Thresholds were selected by fitting a Gaussian to the distribution of {\em Different-ID} pair scores and identifying the score values corresponding to each desired rate. These thresholds were then applied to make a same/different classification on the biometric similarity scores from: (1) the {\em Same-ID} pairs, and (2) the combined ChatGPT and Gemini unmaskings of the {\em US-Senator} dataset. For each condition and threshold, we report the non-match rate: the proportion of faces from the same person that are misclassified as different.


%
%
\begin{figure}[p]
	\centering
	\includegraphics[width=0.85\linewidth]{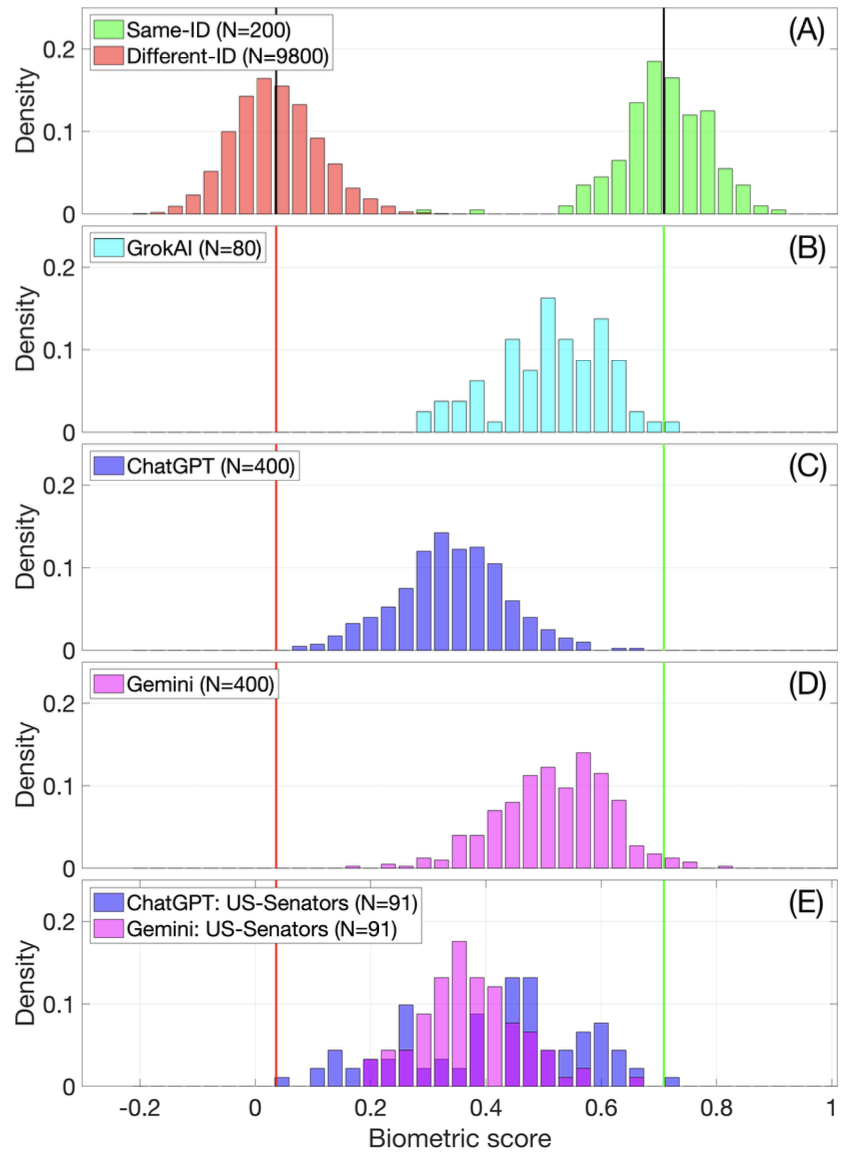}
	\caption{Distributions of biometric facial similarity scores: (A) pairs of images of the same identity (green), images of different identities from the same racial/gender group (red), and images of Doppelgangers (yellow); (B)-(D) pairs of a reference and unmasked image in which the lower-half was removed; and (E) pairs of a reference image and an image in which COVID masks were removed.}
	\label{fig:densities}
\end{figure}

\section*{Results}

\subsection*{AI-powered unmasking produces relatively poor facial fidelity}

Before evaluating AI-powered unmasking, we establish calibration baselines by computing biometric similarity for pairs of images of the same person and pairs of different people within the same gender and race category (Figure~\ref{fig:densities}(A)). The mean (std.dev.) similarity score for the same identities (green) is $0.71$ $(0.08)$, and $0.04$ $(0.08)$ for the different identities (red). Illustrated in Figure~\ref{fig:densities}(B)-(D) are the same distributions for the unmasking of the faces in the {\em Different-ID} dataset using (B) GrokAI, (C) ChatGPT, and (D) Gemini (example unmaskings are shown in the bottom row of Figure~\ref{fig:original-unmasked}). The red and green vertical lines in each panel corresponds to the mean of the different identity (red) and same identity (green) distributions. The mean (std.dev.) of the ChatGPT unmaskings are $0.34$ $(0.10)$, Gemini $0.52$ $(0.10)$, and GrokAI $0.51$ $(0.10)$. By this measure of facial similarity, Gemini and GrokAI reconstruct faces closer to the original than ChatGPT, but all three generate biometric matches notably lower than what is expected for the same identity (Cohen's D: $2.3$ (GrokAI), $4.1$ (ChatGPT), $2.1$ (Gemini)). Similar results are obtained using the AdaFace similarity metric (Cohen's D: $2.1$ (GrokAI), $3.8$ (ChatGPT), $1.8$ (Gemini)).

To determine if there are any differences across the included genders and races, we performed a pair of 2-way ANOVAs (separately for ChatGPT and Gemini). For both ChatGPT and Gemini, there is a significant main effect of gender, with overall lower biometric similarity scores for unmasked females (ChatGPT: F(1,392)=$14.2$, p=$0.0002$; Gemini: F(1,392)=$14.6$, p=$0.0002$). There is also a significant main effect for race (ChatGPT: F(3,392)=$10.3$, p$<0.0001$; Gemini: F(3,392)=$4.1$, p$<0.007$). Follow up comparisons using the Tukey method show that, for ChatGPT, South Asian faces lead to higher biometric scores compared to other races (all mean differences $[0.45,0.69]$, all ps $< 0.005$), and for Gemini, East Asian faces lead to lower biometric scores compared to other races (all mean differences $[0.038,0.039]$, all ps $< 0.03$). These demographic differences suggest that AI-powered unmasking models do not perform uniformly across gender and race, and that conclusions about reliability drawn from one demographic group may not generalize to others. Whether these differences reflect biases in model training data, differences in the information available in the upper half of the face across demographic groups, or other factors remains an open question.

The results from these unmasked images should be interpreted with two caveats. First, biometric similarity for the simulated unmasking was calculated for pairs of images with matched head pose, lighting, and environment (Figure~\ref{fig:original-unmasked}), which may inflate similarity scores relative to real-world conditions. On the other hand, the complete removal of the lower half of the face represents a particularly challenging reconstruction scenario, since all information about the lower facial structure is discarded; a well-fit mask, by contrast, may preserve some structural information about the face beneath it, potentially leading to higher similarity scores than we observe here. To address these two issues, we next investigated facial similarity after removing masks from the $91$ faces in the {\em US-Senators} dataset using ChatGPT and Gemini, for which we had pairs of masked and unmasked images taken at different times/contexts. The resulting distribution of biometric scores is shown in Figure~\ref{fig:densities}(E), squarely occupying the valley between the different and same distributions (mean (std.dev.): ChatGPT, $0.41$ $(0.16)$; Gemini, $0.38$ $(0.09)$). We note that, in addition to the limited demographic diversity in this dataset, reconstruction quality for named public figures may be influenced by the frequency with which their images appear in model training data. If this effect is present, however, it would only strengthen our central conclusion that AI-powered unmasking is biometrically unreliable.

Generative-AI models are non-deterministic, meaning that each time a model is prompted to generate an image, it will produce a different result. To understand this variability, using ChatGPT we generated $100$ unmasked images of one of the US Senators. As before, each unmasked face was biometrically scored against the Senator's reference face. These scores ranged from a minimum of $0.29$ to a maximum of $0.54$, with a mean (std.dev.) of $0.42$ $(0.05)$. Results from Gemini are slightly less variable with a minimum biometric similarity score of $0.21$ to a maximum of $0.36$, with a mean (std.dev.) of $0.28$ $(0.04)$. There is also variability across models. The squared Pearson correlation coefficient between ChatGPT and Gemini scores of the US Senators is $0.18$. These ranges suggest high variability in model performance and further emphasize the unsuitability of these tools for use in unmasking for biometric identification. 

\subsection*{AI-powered unmasking produces high error rates in operational face matching}

Based on these results, we expected that any realistic deployment of AI-powered unmasking would produce an unacceptably high identification failure rate. We therefore evaluated AI-powered unmasking in an operational context. A set of thresholds on biometric similarity were selected to achieve different false match rates. Illustrated in Table~\ref{tab:facematching} are the resulting non-match rates for both the Same-ID condition and the AI-unmasked Senator faces (the percentage of image pairs of the same person that were incorrectly classified as different people). For the Same-ID pairs, non-match rates are negligible across all thresholds, as expected for genuine same-identity pairs. By contrast, AI-unmasked faces yield dramatically elevated non-match rates: even at the most permissive threshold ($1$ in $100$ false match rate), between $6\%$ and $8\%$ of unmasked faces fail to match their reference, rising to between $45\%$ and $52\%$ at the most stringent threshold ($1$ in $1,000,000$). 

\begin{table}[t]
	\begin{center}
			\begin{tabular}{r|cc|cc}
				& \multicolumn{2}{c}{non match} & \multicolumn{2}{c}{non match} \\
				false match & Same-ID & AI-unmasked & Same-ID & AI-unmasked \\
				\hline
				1.0\%    &  0.0000\% & 8.1\%     &  0.0000\% & 6.3\% \\
				0.1\%    &  0.0000\% & 17.3\%    &  0.0000\% & 13.7\% \\
				0.01\%   &  0.0001\% & 28.6\%    &  0.0000\% & 23.1\% \\
				0.001\%  &  0.0006\% & 40.7\%    &  0.0001\% & 33.8\% \\
				0.0001\% &  0.0049\% & 52.3\%    &  0.0012\% & 44.5\% \\
				\hline
				& \multicolumn{2}{c|}{{\bf ArcFace}} & \multicolumn{2}{c}{{\bf AdaFace}} \\
		\end{tabular}
	\end{center}
	\caption{Face detection accuracy for a set of fixed false match rates (incorrectly identifying different identities as same). For each false match rate, non match rates are reported: the percentage of image pairs of the same identity incorrectly identified as different. The ``Same-ID'' column corresponds to the Same-ID dataset, and the ``AI-unmasked'' column corresponds to the combined ChatGPT and Gemini unmasking of the Senators dataset (see  Figure~\ref{fig:densities}(A) and (E)).}
	\label{tab:facematching}
\end{table}
%
%


\section*{Discussion}

Generative AI is capable of producing compelling photo-realistic images which may create the illusion that the resulting images reflect reality. We conclude, however, that AI-powered unmasking is not suitable for biometric identification. We note that our findings are complementary to, and distinct from, work on mask-aware face recognition systems, which match masked faces against known reference databases~\cite{ngan2022}. The scenarios we study, in which the subject's identity is entirely unknown, are not amenable to such systems, and it is precisely in these cases that generative AI unmasking tools are being deployed. 

It might be argued that as generative-AI models continue to improve, they may improve in their ability to reconstruct veridical facial features. But even custom-purpose techniques to reconstruct occluded faces struggle to produce reliable identifications~\cite{din2020novel,li2024recovery}, suggesting that the fundamental challenge lies not in the power of the model but in the nature of the problem: today's AI models are based on statistical inference, and therefore an unmasked image is at best a good guess about the likely facial appearance. Such guesses cannot, in their current form, be applicable for biometric identification.


\bibliographystyle{unsrtnat}
\bibliography{main} 


\end{document}